\documentclass[review]{elsarticle}
\usepackage{tabu}
\usepackage{subcaption}
\usepackage{lineno}
\linenumbers

\journal{Journal of \LaTeX\ Templates}

\biboptions{authoryear}

\usepackage{hyperref}
\hypersetup{pdfauthor={Name}}

\begin{document}

\nolinenumbers

\begin{frontmatter}

\title{Supervised machine learning techniques for data matching based on similarity metrics}

\author[mymainaddress]{Pim Verschuuren\corref{mycorrespondingauthor}}
\cortext[mycorrespondingauthor]{Corresponding author}
\ead{pim.verschuuren@rhul.ac.uk}

\author[mysecondaryaddress]{Serena Palazzo}
\ead{serena.palazzo@ed.ac.uk}

\author[mythirdaddress]{Tom Powell}
\ead{tpowell@fiscaltec.com}

\author[mythirdaddress]{Steve Sutton}
\ead{ssutton@fiscaltec.com}

\author[mythirdaddress]{Alfred Pilgrim}
\ead{apilgrim@fiscaltec.com}

\author[mysecondaryaddress]{Michele Faucci Giannelli}
\ead{michele.faucci.giannelli@ed.ac.uk}

\address[mymainaddress]{Royal Holloway, University of London, Egham, TW20 0EX, U.K.}
\address[mysecondaryaddress]{University of Edinburgh, Old College, South Bridge, Edinburgh EH8 9YL, U.K.}
\address[mythirdaddress]{FISCAL Technologies Ltd, 448 Basingstoke Road, Reading, RG2 0LP, U.K.}

\begin{abstract}
Businesses, governmental bodies and NGO's have an ever-increasing amount of data at their disposal from which they try to extract valuable information. Often, this needs to be done not only accurately but also within a short time frame. Clean and consistent data is therefore crucial. Data matching is the field that tries to identify instances in data that refer to the same real-world entity. In this study, machine learning techniques are combined with string similarity functions to the field of data matching. A dataset of invoices from a variety of businesses and organizations was preprocessed with a grouping scheme to reduce pair dimensionality and a set of similarity functions was used to quantify similarity between invoice pairs. The resulting invoice pair dataset was then used to train and validate a neural network and a boosted decision tree. The performance was compared with a solution from FISCAL Technologies as a benchmark against currently available deduplication solutions. Both the neural network and boosted decision tree showed equal to better performance.\\
Code repository: \url{https://github.com/pimverschuuren/Deduplication}
\end{abstract}

\end{frontmatter}

\section{Introduction}
\noindent
With the rise of the information age in the past decades, the correct handling of data has become a necessity throughout society. However, many systems and processes that integrate and store data are error prone due to human factors, errors in the digitisation and poor system integration. The account payable department of any company is an area that is particularly affected by these problems since it has to handle invoices coming from multiple sources and add them to a single system. The result can be the duplication of an invoice record caused by subtle and small differences in the fields that define the invoice record. The field that deals with these issues is over five decades old [\cite{Newcombe1959,MaximumLinkage,ModelLinkage,TheoryLinkage,HandbookLinkage}] and is known as data matching, record linkage, entity resolution or field matching. Many of the available solutions are rule-based models devised by domain experts as proposed by Wang and Madnick~[\cite{RuleBased}]. In this study a framework is presented that uses domain specific knowledge, combines it with distance-based similarity functions to generalise invoice comparison and identify the duplicates with supervised learning tools. These tools have shown to be very good in finding hidden patterns and correlations in data which can be exploited to categorise the data in different groups. Previous studies [\cite{prevFirst,prevSVMtrees,prevDisclosure,prevConv,prevDeep}] have already shown that supervised machine learning techniques can be fruitful for duplication detection. The value of this study is a new comparison between machine learning techniques and a rule-based solution from industry with a dataset of invoice pairs that is realistic in both volume and the ratio of non-duplicates to duplicates, also known as class imbalance. The results show interesting possibilities for the field of data matching.\\
The paper is structured as follows; in Section 2 the data and its pre-processing is described. Section 3 contains a brief summary on the chosen supervised machine learning techniques and their specific architecture in this application. In Section 4 the training and evaluation of the models is described. The conclusions and outlook on possible future studies are given in Section 5.

\section{The Dataset}
\noindent
A set of invoice records is used as dataset for the study described in this article. The dataset was provided by FISCAL Technologies and sampled from various industries. Each invoice consists of several fields eg. invoice number, invoice date, supplier) that can have text and/or number based values. In order to have a clean training dataset which allows for better performances, invoices with missing information were removed. Note that the removal of invoices is valid as long as the removal percentage w.r.t. the whole dataset remains low~(eg. $<1\%$ in this study). After data cleaning, the pairing step is performed. Pairing requires each invoice to be compared with all the remaining invoices which for $n$ invoices would give $n(n-1)/2$ unique invoice pairs. Given the large number of invoices received by FISCAL Technologies (up to $10^{6}$ per company) this is not computationally feasible. Furthermore, this naive pairing approach would be very inefficient as certain invoices would clearly not match. To solve this issue, a procedure called \textit{blocking}~[\cite{blocking1,blocking2,Blocking3,Blocking4}] is used to reduce the dimensionality of the matching process. Duplicate invoices are very likely to have at least one shared field value. By grouping invoices that have certain field values in common, it is possible to split the dataset into non-disjoint subsets of invoices. Each invoice is then only further compared with invoices from within that subset. This reduces the number of pairings and significantly speeds up the process. Comparing invoices is non-trivial due to the variety of field types that makes an invoice. Fields involving dates, plain numbers, currency and text fields not only have different structures and meanings but their errors may also originate from different sources. Hence, several dedicated similarity algorithms were defined to compare different fields in several ways. All the similarity algorithms take two values as input and try to quantify the degree of similarity with their own unique approach. All of the outputs were normalised such that 1 indicates an exact match and 0 indicates complete dissimilarity. The considered algorithms are listed below.\\
\\
\noindent
\textit{Jaro and Jaro-Winkler}\\
The Jaro similarity algorithm was developed by M. Jaro in [\cite{jaro}]. It was initially designed to compare short length strings such as names. This algorithm was extended by W. Winkler et al. in~[\cite{winkler}] where strings that matched in the beginning get higher similarity scores.\\
\\
\noindent
\textit{N-gram}\\
The N-gram similarity algorithm was proposed by G. Kondrak in~[\cite{ngram}]. This algorithm compares contiguous sequences of $n$ characters also known as $n$-grams. Only 2-, 3- and 4-grams were used in this work. \\
\\
\noindent
\textit{Smith-Waterman}\\
The Smith-Waterman algorithm was developed for DNA-sequencing as proposed in~[\cite{Smith}]. The algorithm tries to find matching character sequences with dynamic programming, a programming approach that deconstructs problems into simpler subproblems.\\
\\
\noindent
\textit{Levenshtein and Damerau-Levenshtein}\\
The Levenshtein algorithm~[\cite{levenshtein}] tries to find the number of single character operations (eg. insertion, deletion or substitution) needed to change one string into the other. Damerau proposed an extension~[\cite{Damerau}] were a transposition between two adjacent characters was also considered as an edit operation.\\
\\
\noindent
\textit{Longest Common Substring}\\
The longest common substring algorithm was proposed by G. Benson et al. in~[\cite{lcs}]. It tries to find the longest string that is a substring of both compared strings.\\

\noindent
\textit{Binary comparison}\\
The binary comparison compares the two field values and gives a 1 if they are completely the same and a 0 if at least 1 character is different.\\
\\
\noindent
\textit{Monge-Elkan}\\
The Monge-Elkan algorithm~[\cite{MonElkan}] can be applied to a string that is constructed from multiple strings divided by spaces eg. a sentence. It takes all possible combinations of the shorter strings and applies one of the before mentioned similarity algorithms on each combination. The average of these similarity scores is then passed as the final  similarity score.\\

\begin{figure}[h!]
    \centering
    \includegraphics[height=13cm]{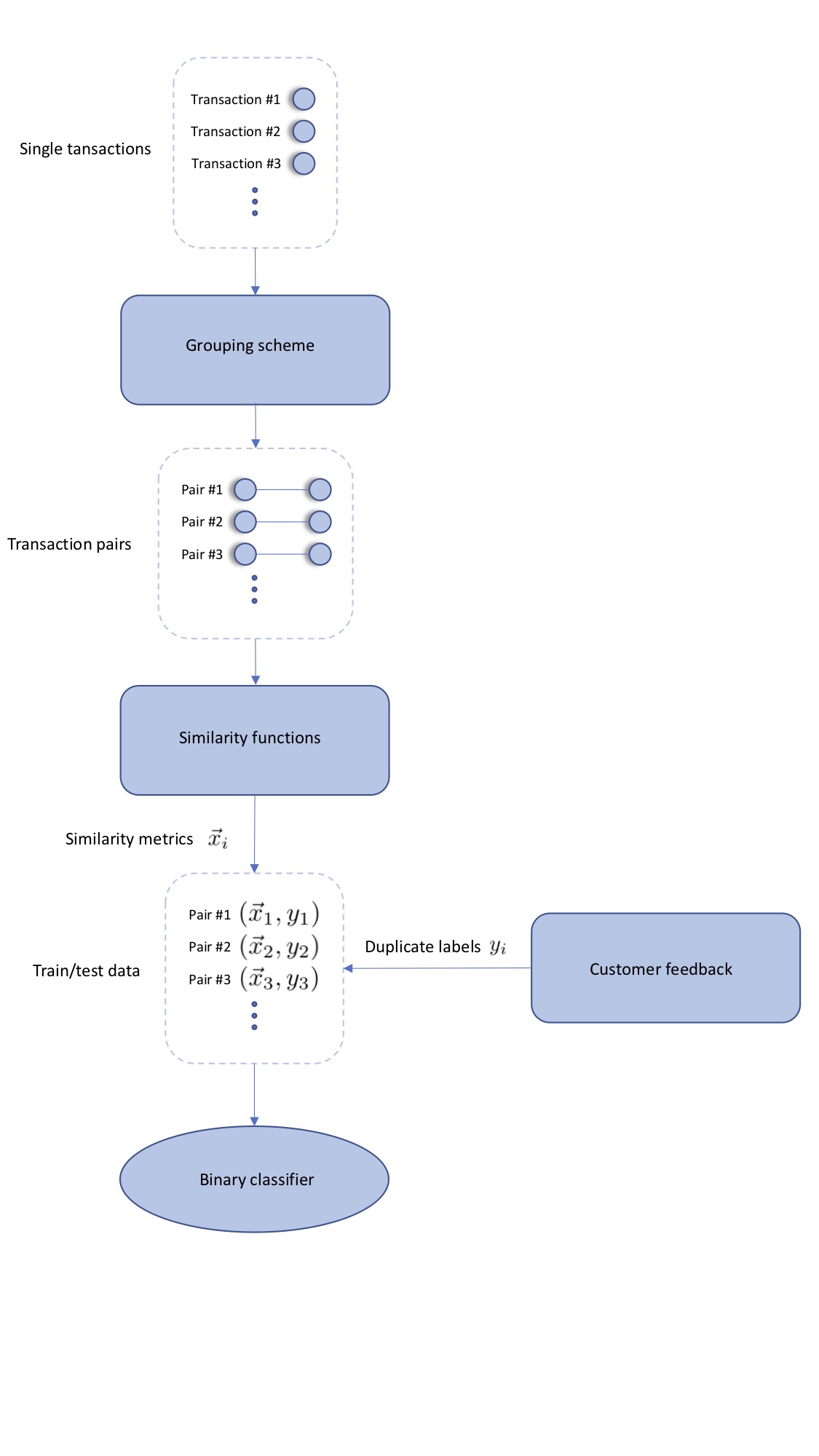}
    \caption{Dataset creation flow chart}
    \label{fig:flowchart}
\end{figure}

\noindent
The similarity algorithms for fields involving strings are well established but algorithms for purely numerical fields~(value, time and age) are still quite underdeveloped. The above set of algorithms has been selected such that as many different field types as possible were covered. The result is a vector of similarity scores for each invoice pair that will be used as input for the machine learning algorithms. All of the invoice pairs are then labelled either duplicate or non-duplicate with the use of customer feedback. An overview of the data and preprocessing workflow is given in figure~\ref{fig:flowchart}.

\section{Machine Learning Approaches}
\noindent
Deduplication of invoices is a binary classification problem and the labelled invoice pairs makes it possible to use supervised learning techniques. In this setting, the following two architectures were chosen.

\subsection{Boosted Decision Tree}
\noindent
\textit{Boosting}~[\cite{FreudBoosting}] is a method that can be applied to many machine learning algorithms. It takes multiple weak learners, eg. classical decision trees, and combines them into one strong learner. The weak learners are trained sequentially with a training sample that is weighted according to the accuracy of the previous weak learner. Boosted decision trees~(BDTs) are well known for their accuracy and being less prone to overfitting. In this application, an improved form of boosting was used known as \textit{gradient boosting}~[\cite{BoostGrad}]. In this approach each learner does not only fit to the reweighted training data but also to the residuals of the previous learners, the difference between predicted and target value. This addition results in a more stable and faster fit convergence. A set of 200 weak learners were trained with a modified least-squares fitting criterion~[\cite{BoostGrad}], a learning rate of 0.1 and a maximum node depth of 4. The number of weak learners, learning rate and maximum node depth were all simultaneously determined with a cross validation hyper parameter grid search. The gradient boosted decision tree uses the \textit{Gradient Boosting Classifier} implementation of the \textit{Scikit-learn} library~[\cite{scikitlearn}].

\subsection{Neural Network}
\noindent
A neural network (NN) is a machine learning algorithm that is inspired by the way the human brain works. It is based on a collection of connected nodes, like human neurons. In a similar way, feature values are passed to the nodes of the input layer which on their turn pass it on to the nodes of the next hidden layer and so on. The input dense layer consists of 20 nodes, matching the number of considered input features. The network has two hidden layers, both having 30 nodes. All layers use the \textit{ReLU} activation function. The output layer consists out of a single node that uses a \textit{Sigmoid} activation to produce a duplication probability. The training is performed with an \textit{Adam} optimization algorithm~[\cite{adam}] and a \textit{binary cross-entropy} loss function.\\
\\
\noindent
To ensure optimal performance, both frameworks were subjected to hyperparameter tuning with a $5$-fold cross-validation grid-scan~[\cite{HyperScan,CrossValHyp}] on the dataset with the non-duplicate invoice pairs undersampled~[\cite{undersamp,undersamp2}] to a 1-on-1 ratio with the duplicate invoice pairs.

\section{Results}
\noindent
In this section it is shown how the models are trained and validated on the constructed dataset. Two different strategies are presented to fit the models and estimate their performance errors with the dataset at hand.

\subsection{5-fold cross-validation}
\begin{figure}[h!]
\centering
\begin{subfigure}{.5\textwidth}
  \centering
  \includegraphics[width=1.1\linewidth]{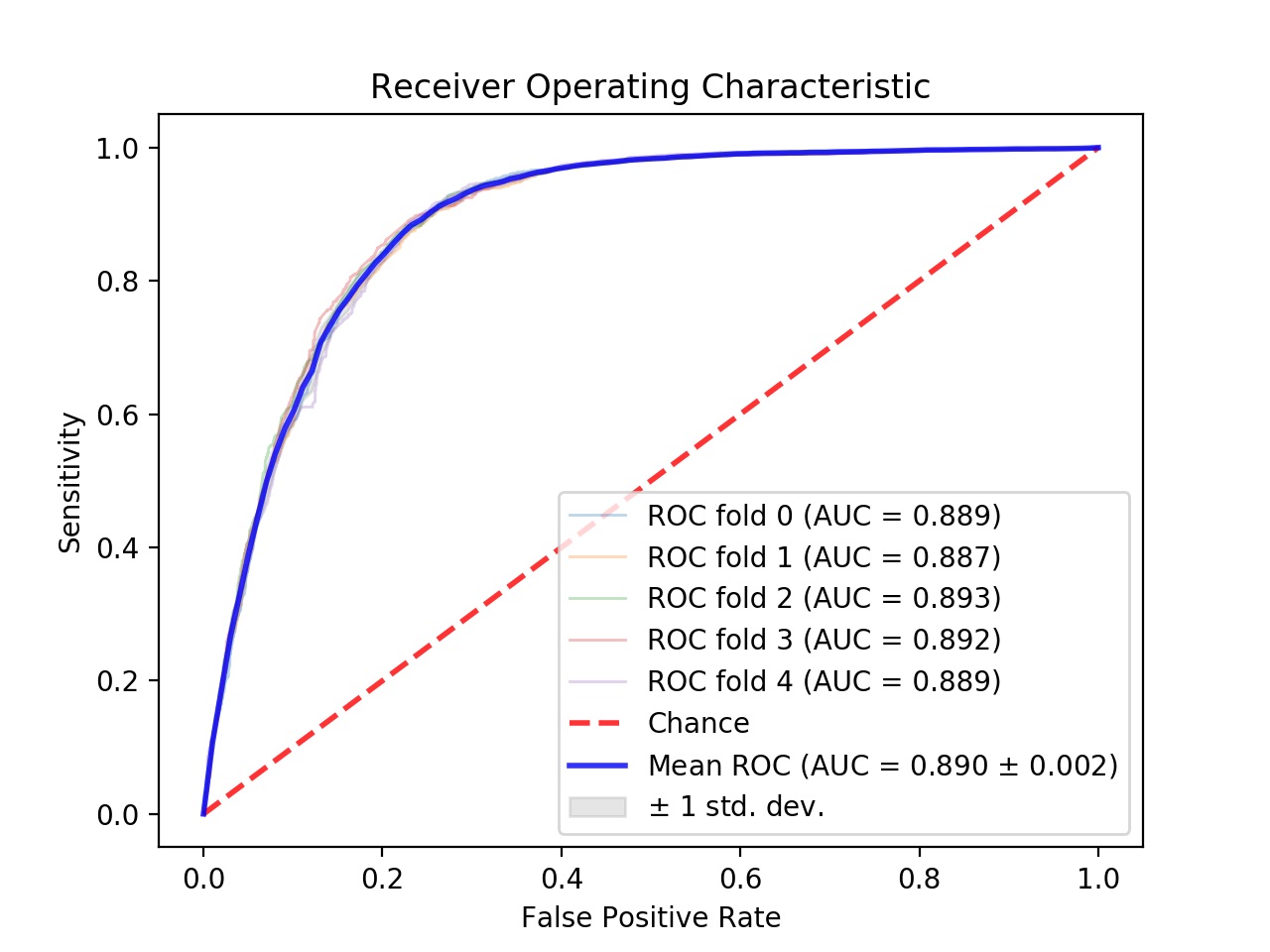}
  \caption{NN}
  \label{fig:sub1}
\end{subfigure}%
\begin{subfigure}{.5\textwidth}
  \centering
  \includegraphics[width=1.1\linewidth]{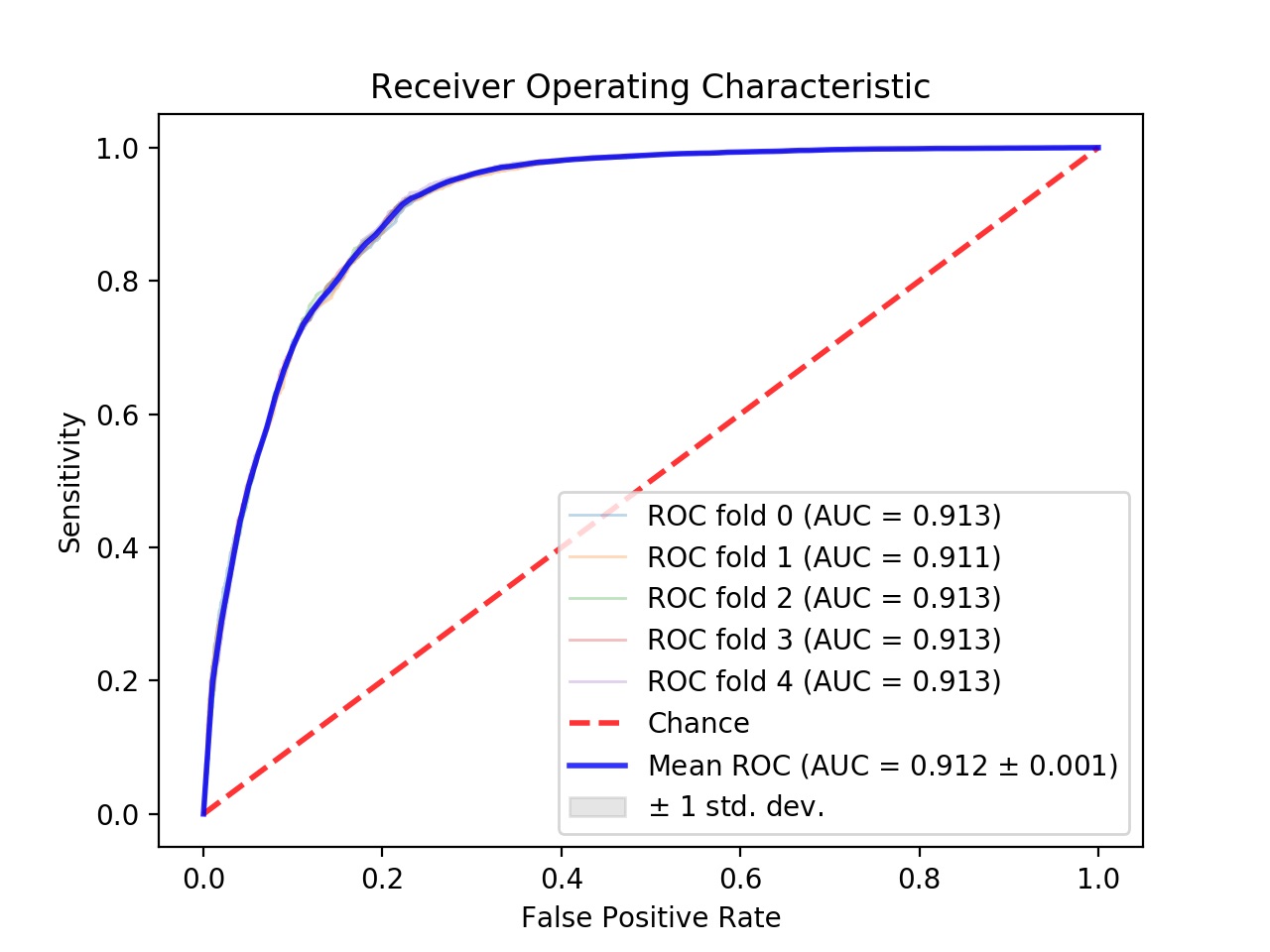}
  \caption{BDT}
  \label{fig:sub2}
\end{subfigure}
\caption{The ROC-curves and AUC's of each folds and the average over all folds of the 5-fold cross-validation}
\label{fig:ROC5fold}
\end{figure}

\noindent
A common model training and validation strategy for a supervised binary classifier is 5-fold cross-validation~[\cite{crossval,crossval2,crossval3}]. The dataset is again undersampled in non-duplicate invoice pairs to a 1-on-1 ratio with the duplicate invoice pairs. The data is then split into 5 even-sized and stratified subsets of which 4 are used for training and 1 for validation. The model is trained and validated 5 times such that each subset is used once for validation. Previous studies have shown that the area-under-curve~(AUC) of a ROC-curve is a good performance metric for the comparison between binary classification models~[\cite{ROC1,ROC2}]. Figure~\ref{fig:ROC5fold} shows that the BDT has a better trade-off between sensitivity~(also known as recall) (Eq. ~\ref{eq:sens}) and false positive rate~(Eq.~\ref{eq:fpr}) than the NN.

\begin{equation}\label{eq:sens}
    \mbox{sensitivity} =  \frac{\mbox{true positives}}{\mbox{true positives + false negatives}}
\end{equation}

\begin{equation}\label{eq:fpr}
    \mbox{false positive rate} = \frac{\mbox{false positives}}{\mbox{false positives + true negatives}}
\end{equation}

\begin{equation}\label{eq:prec}
    \mbox{precision} = \frac{\mbox{true positives}}{\mbox{false positives + true positives}}
\end{equation}

\begin{equation}\label{eq:f}
    F\mbox{-score} = (1-\beta)^{2}\frac{\mbox{precision} \cdot \mbox{recall}}{\beta^{2}\cdot \mbox{precision} + \mbox{recall}}
\end{equation}

\subsection{Imbalanced validation}
\noindent
The second scheme presents the models with training data that is still undersampled but is validated on the data with its original class imbalance of 1 duplicate to a 100 non-duplicates. The invoice pairs are grouped according to the client they originated from.
One client dataset is used in its unbalanced form for validation. The data corresponding to the remaining clients is again undersampled and used as training data. The procedure is repeated until each client has been used once for validation. The predictions of all the clients are then accumulated and evaluated as one testing dataset. The performance is also compared to the predictions of the solution from FISCAL Technologies to provide a benchmark with currently available deduplication frameworks.\\
\\
\noindent
Studies have shown that ROC curves can be misleading in case of binary classification with imbalanced datasets as the dominating true negatives might drive the false positive rate down whilst the precision~(Eq.~\ref{eq:prec}) might still be low. This will result in the invoice pairs identified as duplicates mostly consisting out of false positives. The AUC of the Precision-Recall curve is therefore added as a performance metric as often is advocated for imbalanced datasets~[\cite{PR,PR2,PR3}].\\
\\
\noindent
Moreover, it is relevant to consider that both false positives and false negatives have an impact that affects companies in different ways; the former cause additional work while the latter represent missed duplicates which corresponds to money that cannot be recovered. The relative importance of the two values changes depending on the type of business and the size of the team analysing the invoices. To take this difference of importance into account, it is recommended~[\cite{PR,F1,PR3}] to evaluate the harmonic mean between the precision and recall, also known as the $F$-score~(Eq.~\ref{eq:f}). The $\beta$-parameter changes the $F$-score such that the recall is $\beta$ times more important than the precision. So if a user would determine that the false negatives are 5 times more important than the false positives then one would pick the model with the best $F$-score with $\beta=5$.

\begin{figure}[h!]
\centering
\begin{subfigure}{.5\textwidth}
  \centering
  \includegraphics[width=1.1\linewidth]{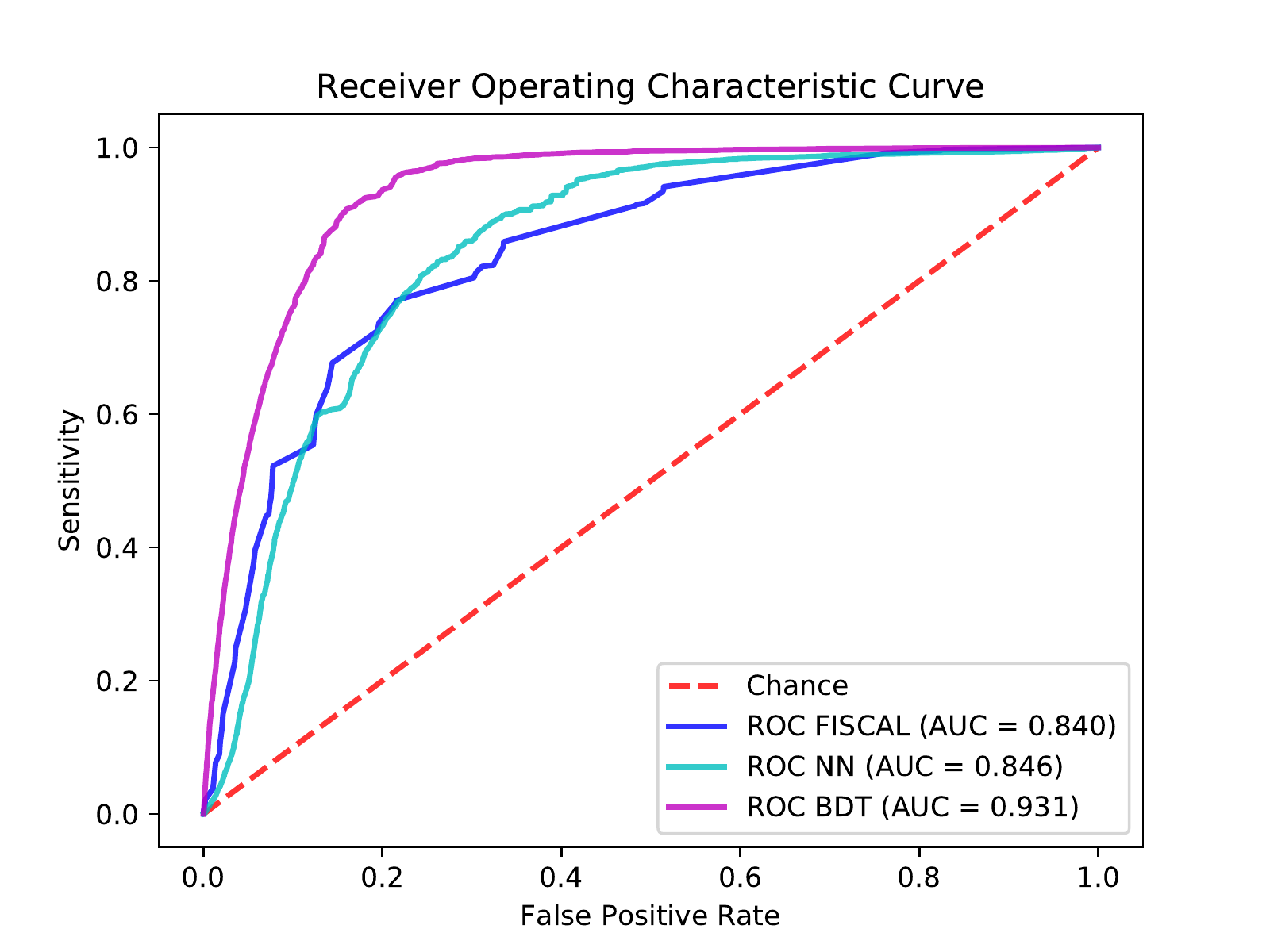}
  \caption{}
  \label{fig:sub1}
\end{subfigure}%
\begin{subfigure}{.5\textwidth}
  \centering
  \includegraphics[width=1.1\linewidth]{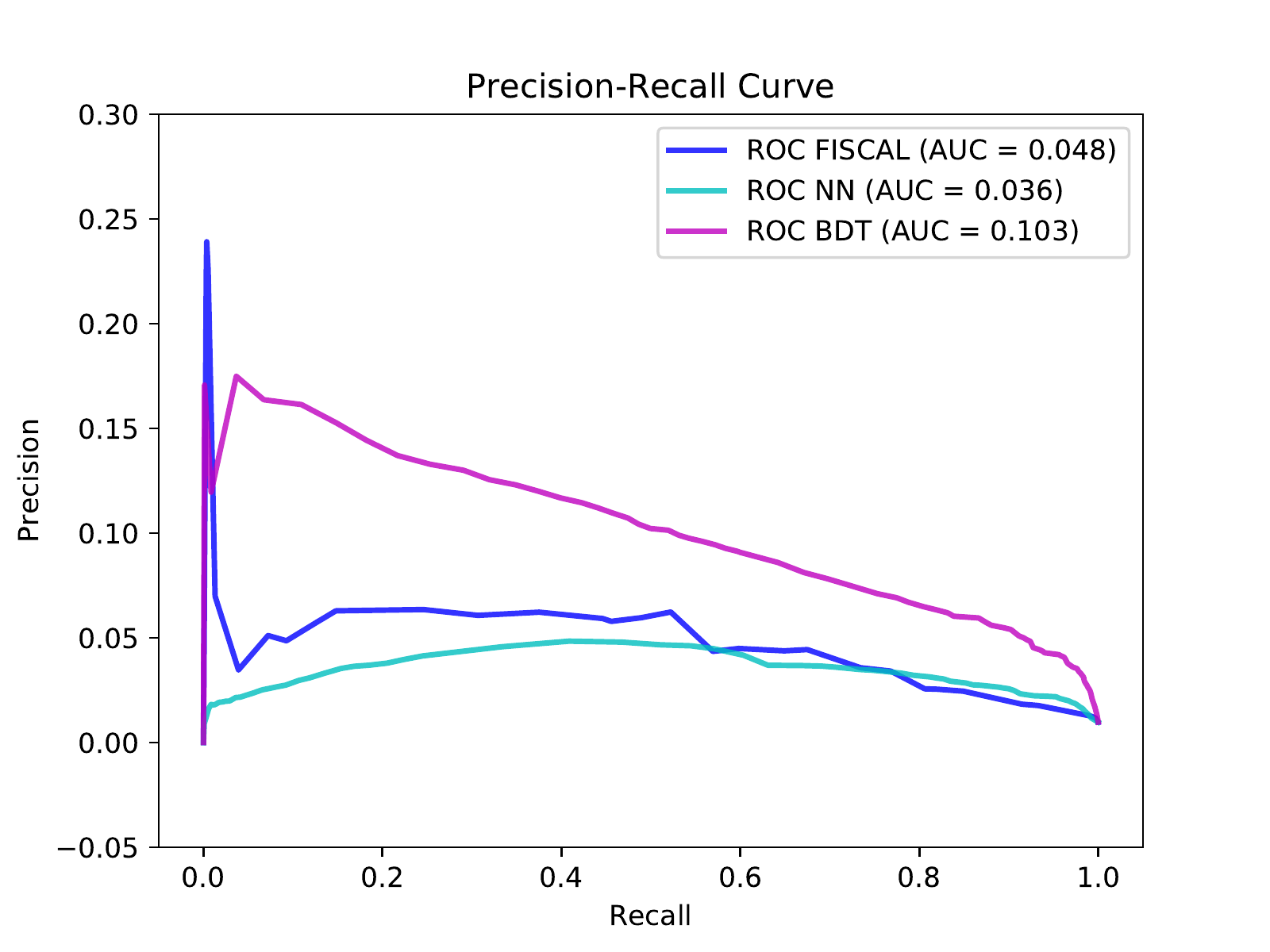}
  \caption{}
  \label{fig:sub2}
\end{subfigure}
\caption{The a) ROC and b) PR curves including AUC's for the imbalanced dataset}
\label{fig:ROCPR}
\end{figure}

\noindent

\begin{figure}[h!]
\centering
\begin{subfigure}{.5\textwidth}
  \centering
  \includegraphics[width=1.1\linewidth]{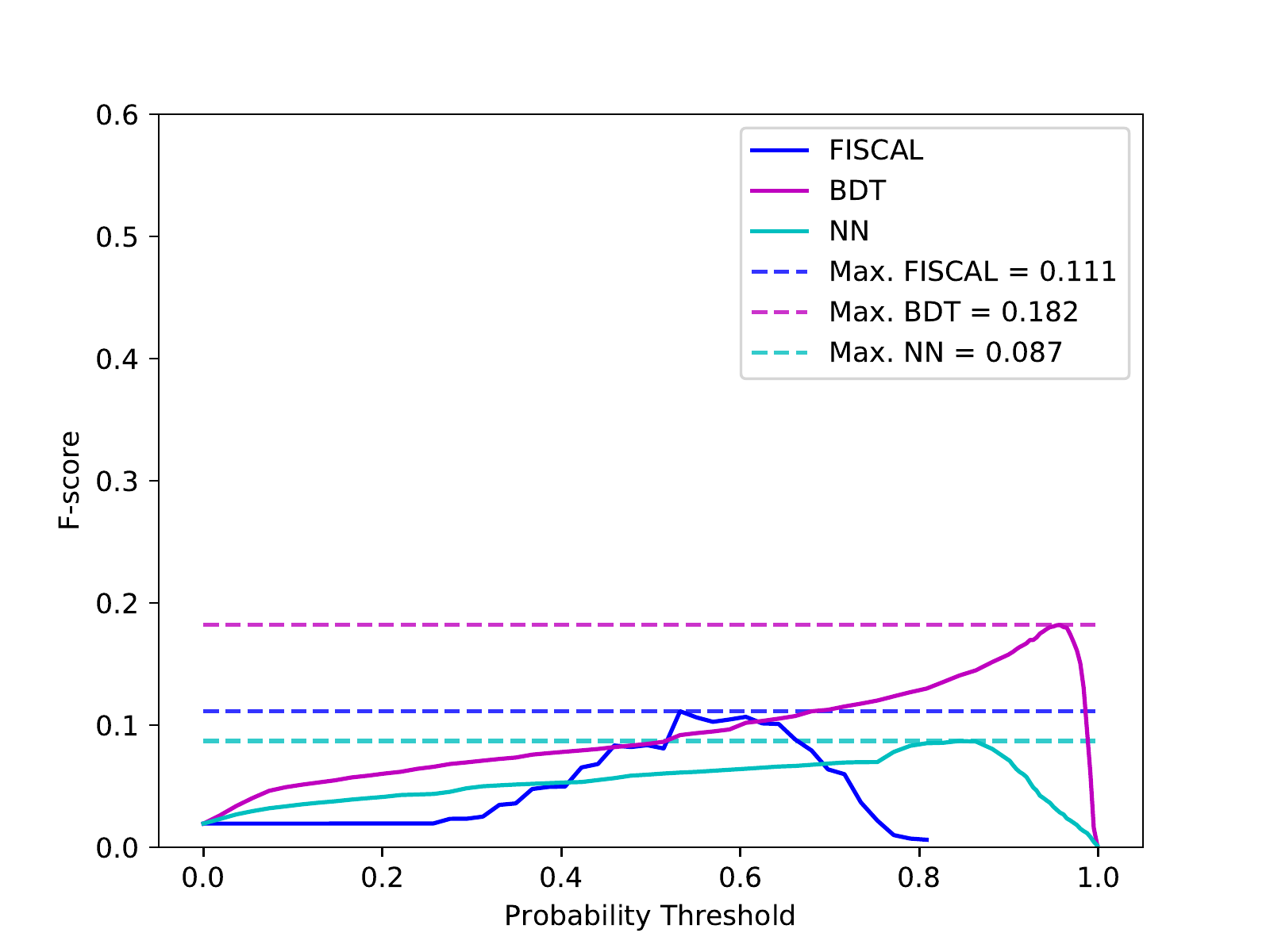}
  \caption{}
  \label{fig:sub1}
\end{subfigure}%
\begin{subfigure}{.5\textwidth}
  \centering
  \includegraphics[width=1.1\linewidth]{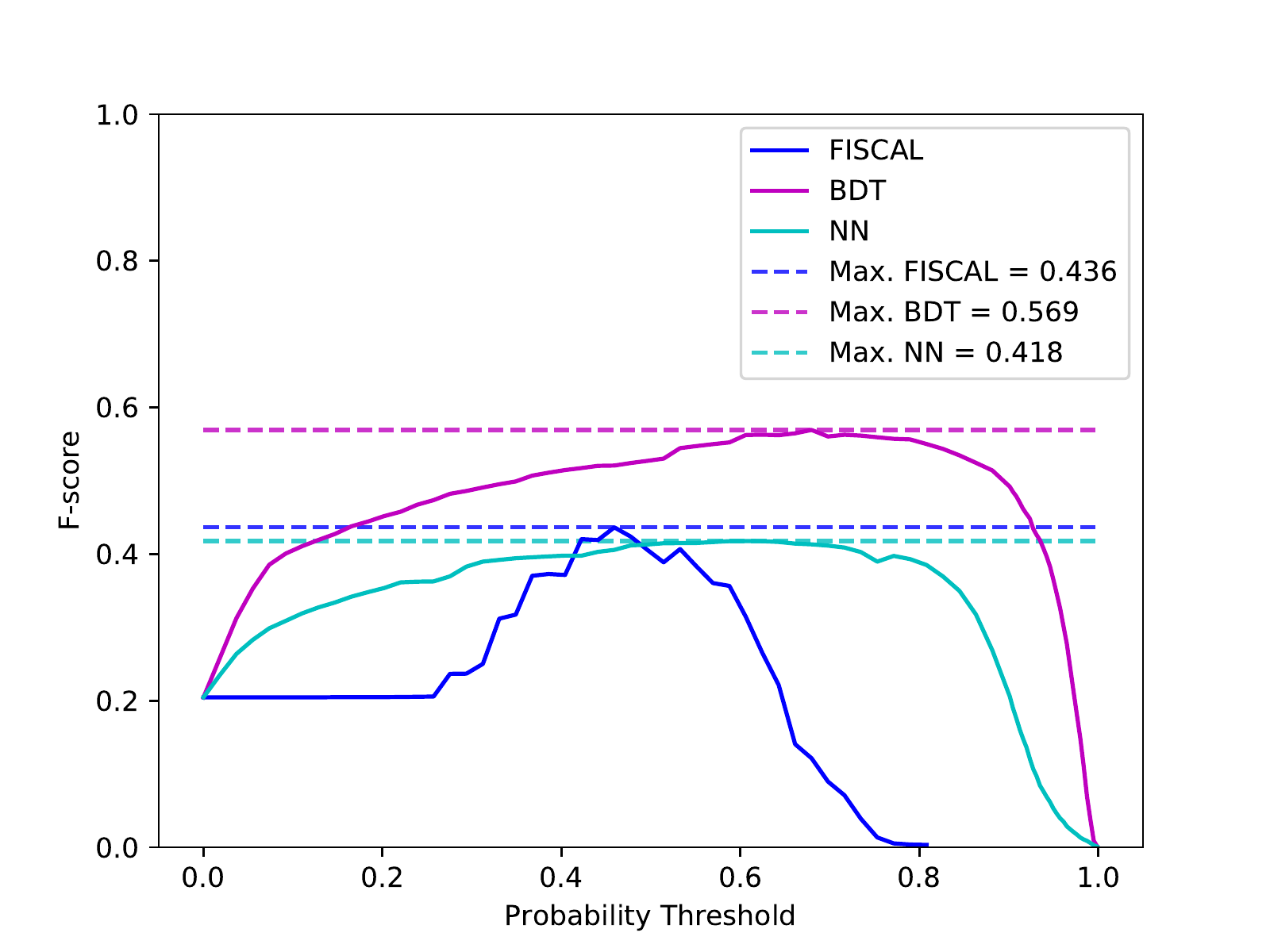}
  \caption{}
  \label{fig:sub2}
\end{subfigure}
\caption{$F$-scores with a) $\beta=1$ and b) $\beta=5$ for all models with varying probability thresholds}
\label{fig:ROCPR}
\end{figure}

\section{Conclusions and outlook}
\noindent
In this work we showed that the identification of duplicated invoices can be significantly improved using supervised learning techniques. A neural network and a boosted decision tree were developed and compared to a market leading solution developed by FISCAL which provided an excellent benchmark. Since each company processes a different fraction of duplicates,  effectively operating at a different probability threshold, it is not possible to quantify the performance of any tool with a single value. Overall, the BDT showed the best performance and could provide a significant improvement to FISCAL for any number of processed invoices. The NN has similar performance to the FISCAL tool but it has the benefit of providing the same  performance for a wider range of probability thresholds, effectively making it a more flexible tool. This could give significant gains to companies operating far from the best working point of the FISCAL tool. \\

\noindent
The presented work also shows promise for future studies. The similarity functions used to construct the features were primarily string based. Adding features that quantify the similarity between numerical fields could add discriminative power to the models. Another addition would be to explore the possibilities of unsupervised learning. Removing the need for~(non-)duplicate labels would substantially increase the size of the dataset at hand. Finally, the class imbalance was handled with undersampling of the training data but remains an issue resulting in a large number of false positives. Research on more advanced methods could further enhance the performance of ML-based tools. 

\clearpage
\section{Acknowledgments}
This research was conducted in partnership with FISCAL Technologies Ltd from their head office in Reading, Berkshire; FISCAL supplied facilities, equipment and test data, as well as the expertise and insight required to ratify the success of the project results.\\

This project has received funding from the European Union Horizon 2020 research and innovation programme under grant agreement No 765710.

\bibliography{mybibfile}

\end{document}